\documentclass[letterpaper, 10 pt, conference]{ieeeconf}  

\IEEEoverridecommandlockouts                              
\usepackage{times}
\usepackage{epsfig}
\usepackage{amsmath}
\usepackage{amssymb}
\newcommand{\netfunction}{f}
\usepackage{booktabs}
\usepackage{multirow}
\usepackage{textcomp}
\usepackage{gensymb}
\usepackage{adjustbox}
\usepackage{soul}

\usepackage{floatrow}
\newfloatcommand{capbtabbox}{table}[][\FBwidth]
\usepackage{graphicx}
\usepackage{comment}
\usepackage{amsmath,amssymb} 
\usepackage{color}
\usepackage{hyperref}

\usepackage[normalem]{ulem}

\usepackage{float}
\floatstyle{plaintop}
\restylefloat{table}

\usepackage{xcolor}

\newcommand{\evin}[1]{\textcolor{black}{#1}}

\newcommand{\etal}{\textit{et al. }}
\newcommand{\ie}{\textit{i.e. }}
\newcommand{\eg}{\textit{e.g. }}

\title{\LARGE \bf
Object-aware Monocular Depth Prediction with Instance Convolutions
}

\author{Enis Simsar$^{1,*}$, Evin P{\i}nar \"{O}rnek$^{1,*}$, Fabian Manhardt$^{2}$, Helisa Dhamo$^{1}$, Nassir Navab$^{1}$ and Federico Tombari$^{1, 2}$
\thanks{$^1$:Technical University of Munich, Germany; {\tt\small \{enis.simsar, evin.oernek, helisa.dhamo, nassir.navab, federico.tombari\}@tum.de}}
\thanks{ $^2$:Google Inc. {\tt\small fabianmanhardt@google.com}}
\thanks{$*$:the first two authors contributed equally.} }

\begin{document}

\maketitle
\thispagestyle{empty}
\pagestyle{empty}

\begin{abstract}

With the advent of deep learning, estimating depth from a single RGB image has recently received a lot of attention, being capable of empowering many different applications ranging from path planning for robotics to computational cinematography. Nevertheless, while the depth maps are in their entirety fairly reliable, the estimates around object discontinuities are still far from satisfactory. This can be attributed to the fact that the convolutional operator naturally aggregates features across object discontinuities, resulting in smooth transitions rather than clear boundaries. Therefore, in order to circumvent this issue, we propose a novel convolutional operator which is explicitly tailored to avoid feature aggregation of different object parts. In particular, our method is based on estimating per-part depth values by means of super-pixels. The proposed convolutional operator, which we dub "Instance Convolution", then only considers each object part individually on the basis of the estimated super-pixels. Our evaluation with respect to the NYUv2, iBims and \evin{KITTI datasets demonstrate the advantages} of Instance Convolutions over the classical convolution at estimating depth around occlusion boundaries, while producing comparable results elsewhere. Our code is available here \footnote {https://github.com/enisimsar/instance-conv}.

\end{abstract}

\section{INTRODUCTION}

\evin{Monocular depth prediction (MDP)} is a very important field in research due to its wide range of applications in robotics and AR~\cite{almagro2019, tateno2017cvpr, Shih3DP20}. Nevertheless, predicting accurate depth from monocular input is also an inherently ill-posed problem. For the human perceptual system depth perception is a simpler task, as we heavily rely on prior knowledge from the environment. Similarly, deep learning has recently proven to be particularly suited for such problems, as the network is also capable of leveraging visual priors when making a prediction~\cite{saxena2005, hoiem2005}.

\begin{figure}[t!]
\begin{center}
\includegraphics[width=\linewidth]{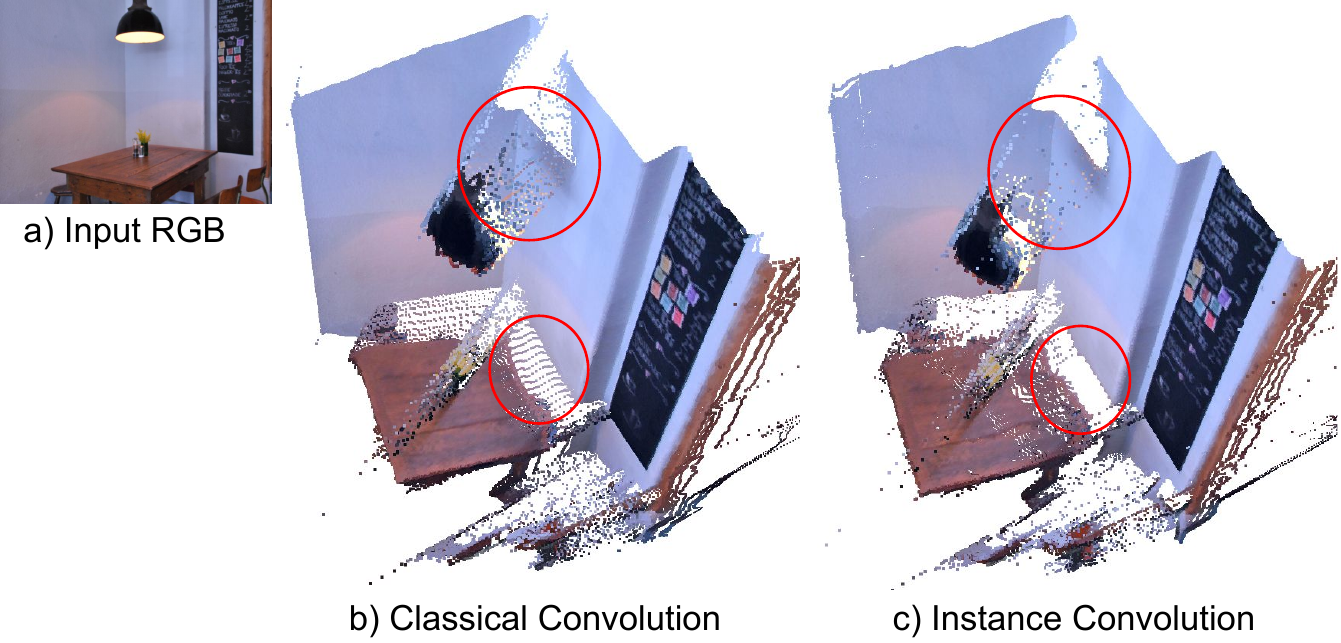}
\end{center}
\caption{
The interpolation effect of classical convolutions induces smeared occlusion boundaries in the predicted depth map (here on an image taken from the iBims \cite{koch2019} dataset), as visible in (b), whereas the proposed Instance Convolution improves on this drawback (c). Note that, while this effect is not as evident when visualizing the 2D depth map, it gets clearly revealed once the depth map is back-projected in 3D.}
\label{fig:intro}
\end{figure} 

With the rise of deep learning and increasing availability of appropriate and large datasets~\cite{silberman2012, koch2019, dai2017scannet, cityscapes2016, Ranftl2020}, depth prediction from single images has recently made a huge leap forward in terms of robustness and accuracy~\cite{wang2020, Miangoleh2021Boosting, yin2019virtualnormal, jiao2018deeper, fu2018dorn}. Yet, despite those large improvements, they still often fall short of adequate quality for specific robotics applications, such as in path planning or robotic interventions where robots need to operate in hazardous environments with low-albedo surfaces and clutter~\cite{almagro2019, koch2019, ramamon2020}. One of the most limiting factors is the poor quality around object edges and surfaces, which directly affects the 3D perception, thereby resulting in a robot missing the objects. The predicted depth maps are typically blurry around object boundaries due to the nature of 2D convolutions and bilinear upsampling. Since the kernel aggregates features across object boundaries, the estimated depth map commonly results in an undesired interpolation between fore- and background. Similarly, associated 3D point clouds cannot reflect 3D structures (see Fig. \ref{fig:intro}). In this work, our motivation is to capture object-based depth values more sharply and completely, while preserving the global consistency with the rest of the scene.

To circumvent the smeared boundary problem, \ie avoid undesired depth interpolation across different segments, we are interested in an operation that extracts features within a continuous object segment. To achieve this, we employ a novel convolution operation inspired by the sparse convolutions introduced by Uhrig \etal~\cite{uhrig2017}. 

Sparse convolutions are characterized by a mask, which defines the region in which the convolution operates. 
While sparse convolutions typically rely on a single mask throughout the entire image, in this work, the mask depends on the pixel location. Given a filter window, we define the mask as the feature region that belongs to the same segment as the central pixel of that window. 
In other words, only the pixels that belong to a certain object contribute to its feature extraction. We name our convolution operator \emph{Instance Convolution}.

Using Instance Convolutions to learn the object depth values should make the depth values at the object edges sharper than regular convolution, \ie prevent the interpolation problem at the occlusion boundaries. Despite this advantage in terms of boundary sharpness, Instance Convolutions come with an obvious drawback. An architecture based solely on such operation, would not be able to capture object extent and global context. This inherent scale-distance ambiguity would thus make it impossible to obtain metric depth. Therefore, we propose an architecture that first extracts global features so to utilize scene priors via a common backbone comprised of regular convolutions. We then append a block composed of Instance Convolutions to rectify the features within an object segment, resulting in sharper depth across occlusion boundaries. Notice that we chose an optimization-based approach~\cite{achanta2012slic}, producing super-pixels, to obtain corresponding segmentation as a deep learning-driven method would simply shift the problem of clear boundaries towards the segmenter. In addition, our method does not require any semantic information, but rather only needs to understand what pixels belong to the same discontinuity-free object part.

Our contributions can be summarized as follows: 

\begin{itemize} 
    \item We propose a novel end-to-end method for depth estimation from monocular data, which explicitly enforces clear object boundaries by means of super-pixels.
    \item To this end we propose a dynamic convolutional operator, \textit{Instance Convolution}, which only aggregates features appertaining to the same segment as the center pixel, with respect to the current kernel location. 
    \item Further, as we are required to properly propagate the correct segment information throughout the whole network, we additionally introduce the "center-pooling" operator, keeping track of center pixel's segment id.
\end{itemize}

We validate the usefulness of Instance Convolutions for edge-aware depth estimation on \evin{three} commonly employed benchmarks, namely NYUv2~\cite{silberman2012},  iBims~\cite{koch2019}, \evin{and KITTI~\cite{kitti}}. Thereby, we show that Instance Convolutions is able to improve object boundaries regardless of the chosen backbone depth estimator.

\section{Related Work}

\paragraph {Supervised monocular depth prediction} The first attempts to tackle monocular depth estimation were proposed by~\cite{saxena2005,saxena2009} via hand-engineered features and Markov Random Fields (MRF). Later, the advancements in deep learning established a new era for depth estimation, starting with Eigen \etal~\cite{eigen2014}.
One of the main problems in learned depth regression occurs in the decoder part. Due to the successive layers of convolution channels in neural networks, fine details of the input images are lost. There are a number of works that approach this problem in different ways. Eigen \& Fergus~\cite{eigen2015} introduced multi-scale networks to make depth predictions at multiple scales. Laina \etal~\cite{laina2016} built upon a ResNet architecture with improved up-sampling blocks to reduce information loss in the decoding phase. Xu \etal~\cite{xu2017} proposed an approach that combines deep learning with conditional random fields (CRF), where CRFs are used to fuse the multi-scale deep features. 

A line of works pursued multitask learning approaches that predict semantic or instance labels~\cite{jiao2018deeper}, depth edges, and normals~\cite{ramamonjisoa2019sharpnet, zhang2019pap, lee2019big} to improve depth prediction. Kendall \etal~\cite{kendall2017multi} investigated the effect of uncertainty estimation in scene understanding. \evin{In contrast, Yin \etal \cite{yin2019virtualnormal} estimate the 3D point cloud from a predicted depth map, using the surface geometry as additional guidance. Recently, Bhat \etal~\cite{Bhat2021AdaBinsDE} proposed a novel formulation for predicting distance values by means of classification. Tian \etal~\cite{Tian2020} proposed attention blocks within the decoder, while other methods adopt a fully Transformer based architecture~\cite{Ranftl2021,yang2021transformers}.}

\paragraph{Occlusion boundaries} All of the above works aim to learn a globally consistent depth map, yet, do not focus on fine local details, often resulting in blurred boundaries and deformed planar surfaces. Consistent with our work, Hu \etal~\cite{hu2018boundary} focuses on accurate occlusion boundaries through gradient and normal based losses. Ramamonjisoa~\etal~\cite{ramamonjisoa2019sharpnet} aims to improve predicted depth boundaries by estimating normals and edges along with depth and establishing consensus between them. In a follow-up work, they apply wavelet decomposition at different scales and re-weight the feature map by a binary mask calculated at different frequency coefficient thresholds~\cite{ramamonjisoa2021}. Several works apply bilateral filters to increase occlusion gaps~\cite{shih2020}, infer each object separately~\cite{dhamo2019}, or learn energy-based image-driven refinement focusing on edges~\cite{niklaus2019, ramamon2020}. \evin{Cliffnet \cite{cliffnet2020}, combines multi-scale features to obtain sharper depth maps, whereas, Tosi \etal~\cite{Tosi2021CVPR} tackles the issue through mixture density networks, focusing on stereo matching problem.}

\begin{figure}[t]
\begin{center}
\includegraphics[width=0.95\linewidth]{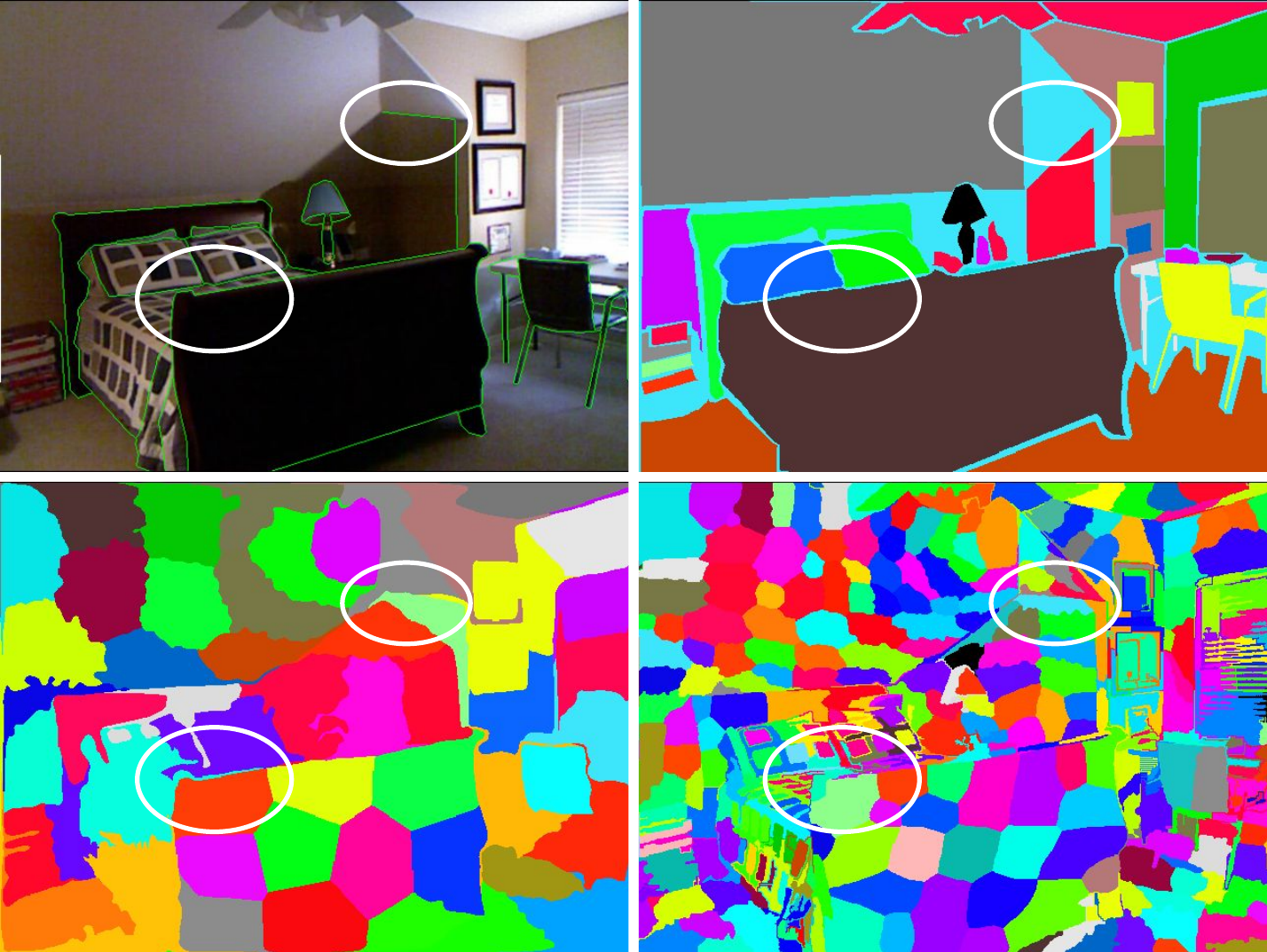}

\end{center}
\caption{\textbf{Different segmentation methods}. The top left shows the input RGB image with occlusion boundaries in green, while the top right illustrates the ground truth segmentation. Notice how the latter does not consider intra-object discontinuities \evin{(highlighted with white-circles)}. Thus, we leverage super-pixels to account for any discontinuities based on the RGB input. Exemplary, we demonstrate the obtained super-pixels for SLIC~\cite{achanta2012slic} using 64 segments on the bottom left, and BASS~\cite{uziel2019bayesian} on the bottom right. Note that BASS outputs approximately 200 segments on average.}
\label{fig:over_seg}
\end{figure}

\begin{figure*}[t!]
\begin{center}
\includegraphics[width=\linewidth]{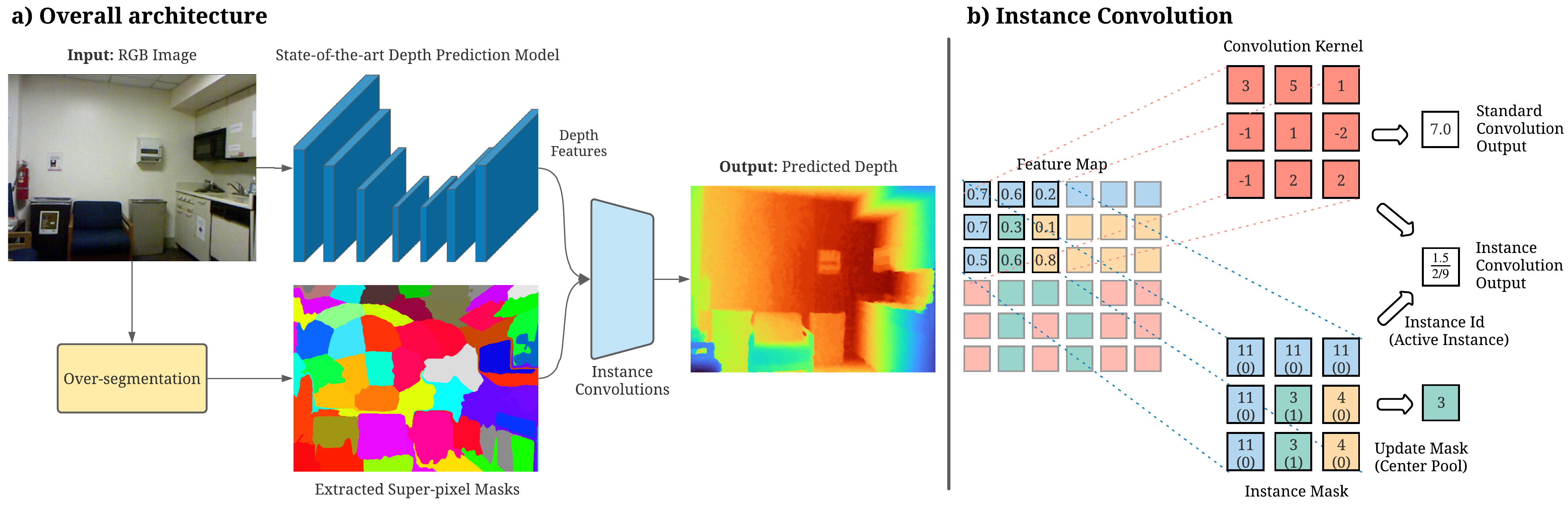}
\end{center}
\caption{\textbf{a) Schematic overview of proposed architecture.} The input image is fed into a state-of-the-art depth prediction model (\eg SharpNet~\cite{ramamonjisoa2019sharpnet} or BTS \cite{lee2019big}) to obtain global image features. The extracted depth features along with the object masks are then passed into the Instance Convolution block to predict a sharp depth map. \textbf{b) Masking mechanism of Instance Convolutions.} The abstract image on the left part contains a chair represented in green. The feature values of the masked region of the kernel are 0.3 and 0.6. The mask for the kernel shows the object pixels. If the normal convolution with a binary mask multiplication is used for this example, 0.9 can be obtained as the result, but with the Instance Convolution, the result is $\frac{0.9}{2/9}$. The successive mask is calculated by taking the center value of the current kernel mask.}
\label{fig:architecture}
\end{figure*}

\paragraph{Sparse convolutions} Sparsity in convolutions has been investigated in several works~\cite{liu2015sparseconv, wen2016nips} aimed at improving the efficiency of neural networks by reducing the number of parameters, \ie increasing sparsity. Minkowski Engine was proposed as an efficient 3D Spatio-Temporal convolution built on sparse convolutions \cite{choy2019minkowski}. In contrast to such works, Uhrig \etal used sparse meshes \cite{uhrig2017} to improve structural understanding in the case of sparse inputs, \eg depth map completion ~\cite{zhao2021adaptive}. Some works used a similar convolution operator, called \textit{partial} or \textit{gated} convolution, for image editing and inpainting tasks~\cite{liu2018partialinpainting, yu2019gatedconv} to discard content-free regions. Su \etal \cite{su2019pixel} applies per-pixel kernel weight adaptation through a Gaussian assumption, omitting the boundaries. In our work, we are also interested in computing convolution only on a subset of input pixels. Differently from these works, our masks do not define a random set or a normally distributed sparse set of pixels. Our masks change dynamically for each pixel position, to extract features within the same \textit{object segment}.

\paragraph{Over-segmentation methods} The proposed instance convolution operation relies on the detection of meaningful object segments in a scene. One alternative would be to identify all objects in a scene, either as annotated instance labels, or via learned segmentation models, \eg Mask R-CNN~\cite{he2017maskrcnn}.
The former requires a heavy amount of annotation for large datasets. The latter requires the objects in the corresponding dataset to match the pre-trained models and can additionally lead to inaccurate edges. To detect objects without pre-trained models and labeled data, in this work, we leverage over-segmentation methods. Among the available methods for over-segmentation \cite{achanta2012slic, uziel2019bayesian, levinshtein2009turbopixels}, in our experiments we mainly focus on super-pixel (SLIC) \cite{achanta2012slic} and Bayesian adaptive super-pixel segmentation (BASS) \cite{uziel2019bayesian} (see Fig. \ref{fig:over_seg}).

\section{Methodology}

In this section, the problem statement and the individual components of the proposed method for boundary-aware MDP are presented.

\subsection{Depth Estimation Using Deep Learning}

Monocular depth estimation has recently received a lot of attention in literature and several methods have been proposed~\cite{wang2020, Miangoleh2021Boosting, Ranftl2020}. Interestingly, even very early methods have noticed the performance degradation around occlusion boundaries, and various different measures, such as skip-connection \cite{eigen2015, laina2016} or Conditional Random Fields~\cite{xu2017, liu2015}, have been put in place to counteract the smearing effect. Nevertheless, despite those measures, the proposed methods still can not capture the high frequencies of object discontinuities due to the inherent nature of 2D convolutions.

The classical convolution kernel simultaneously operates on all inputs within the kernel region, performing a weighted summation. Consequently, features originating from different object parts are simply fused, which in turn causes a blurring of the object boundaries, \ie the edges that separate the object from the background in 3D. A corresponding example is shown in Fig.~\ref{fig:intro}. Notice that this effect is more visible when viewing the associated 3D point cloud.

\subsection{Instance Convolutions for Boundary-Aware Depth Estimation}

To avoid aggregation of features appertaining to different image layers, we thus propose to leverage super-pixels in an effort to guide the convolution operator. In particular, inspired by Sparsity Invariant Convolutions \cite{uhrig2017}, we propose \textit{Instance Convolutions}, which applies the weighted summation only to pixels belonging to the same segment as the central pixel. \evin{In other words, it renormalizes the convolution function by considering the number of instance pixels that contributed to the output to keep it at a similar range than the common convolutional operator, as depicted in Fig. 4. Notice that when all pixels are within the instance mask, our formulation simply turns into standard convolution. Formally, for the n-th output feature map this can be written as}

\evin{
\begin{equation}
\netfunction_{u,v,n} (S, X, W)
 = \frac{
  \sum_i (S_{u,v} \odot X_{u,v} \odot W_n)_i}
 { \frac{ \sum_i {(S_{u,v})_i}}{k_{w}*k_{h}} +\epsilon} + b,
\label{eq:sparse_convolution_full}
\end{equation}
with $\odot$ denoting the element-wise multiplication, $X_{u,v}$ being the features within the window of the kernel of size $k_w \times k_h$ placed at $(u,v)$, $S_{u,v}$ being the indicator kernel which is $1$ for each pixel within the window if it belongs to the same class as the central pixel $(u, v)$ and 0 otherwise, and $W_n$ denoting the applied kernel weights. All matrices $S_{u,v}, X_{u,v}, W_{n} \in \mathrm{N}^{d_{in}\times k_w \times k_h}$ with the indicator kernel $S_{u,v}$ being replicated, accordingly.
}

Since our architecture follows the standard encoder-decoder methodology (Section~\ref{sec:arch}), we have to adequately propagate the segment information through the network. However, as MaxPooling can lead to loss of spatial information, we introduce the center-pooling operator, which simply forwards the segment id of the central pixel with respect to each downsampling operation to preserve the object boundaries. Whereas at upsampling, the original semantic map (or downsampled from previous layers) is directly used as they are already readily available or computed. For a detailed explanation of Instance Convolutions, see Fig. \ref{fig:architecture} (b).

A deep learning-based approach would simply transfer the problem of clear boundaries towards the segmenter. Such methods can also never cope with all objects classes in the wild. Moreover, our method does not require any semantic information, but rather only needs to understand what pixels belong to the same discontinuity-free object part. Hence, we instead rely on optimization-based approaches, \ie SLIC~\cite{achanta2012slic} and BASS~\cite{uziel2019bayesian}, to obtain the needed super-pixels. Thereby, these works can provide not only object boundaries but also self-occlusions within objects (see Fig.~\ref{fig:over_seg}). Noteworthy, while SLIC requires to define the number of output segments in advance, BASS can find the optimal number of segments by itself, however at a higher computational cost. 

\subsection{End-to-end Architecture}\label{sec:arch}

To summarize, we model our Instance Convolution such that the method is particularly suited for estimating depth at object boundaries. Nevertheless, as an output pixel has never observed any feature outside of the segment it resides on, it is impossible for the model to predict metric depth due to the scale-distance ambiguity (\ie a large object far away can have the same projection onto the image plane as a small object close by). Therefore, we harness a state-of-the-art MDP backbone to extract global information about the scene. We then feed the extracted features together with the obtained super-pixels to our Instance Convolution-driven network to estimate the final edge-aware depth. Since the backbone as well as our Instance Convolution block are fully differentiable, we can train the whole model end-to-end. Proposed method can be plugged together with different depth predictors. In this paper, we use SharpNet \cite{ramamonjisoa2019sharpnet}, BTS \cite{lee2019big}, and VNL \cite{yin2019virtualnormal} for feature extraction, to show the generalizability of Instance Convolutions. 

\subsection{Training Loss}
Our training objective is composed of three terms, \ie the depth loss $L_{1}$, the gradient loss $L_{grad}$, and the normal loss $L_{normal}$. \evin{Notably, these are standard choices also used by \cite{hu2018boundary, chen2019structure}.  The $L_{1}$-term is the main cost function to guide the learning of per-pixel distance values by taking the absolute difference between the predicted depth $d$ and the ground truth depth $d^{GT}$} with:

\begin{equation}
    L_{1}(d, d^{GT}) = \frac{1}{N} \sum_{i=1}^{N} | d^{GT}_{i} - d_{i} |.
\end{equation}

\noindent Furthermore, the depth gradient loss is given as
\begin{equation}
    L_{grad}(d, d^{GT}) = \frac{1}{N} \sum_{i=1}^{N} |\nabla_{h}d_{i} - \nabla_{h}d_{i}^{GT}| + |\nabla_{v}d_{i} - \nabla_{v}d_{i}^{GT}|,
\end{equation}
\noindent \evin{and penalizes missing fine details. The horizontal ($\nabla_{h}$) and vertical ($\nabla_{v}$) gradients are computed via the Sobel operator.}

\evin{Finally, in an effort to learn finer-level surface details, such as high-frequency fluctuations on small structures, we utilize the angular distance between per-pixel normals according to}
\begin{equation}
    L_{normal}(n, n^{GT}) = \frac{1}{N} \sum_{i=1}^{N} \left( 1 - \frac{\langle n_{i},n_{i}^{GT}\rangle}{||n_{i}|| \cdot ||n_{i}^{GT}||}  \right).
\end{equation}
\evin{Thereby, a surface normal $n_{i}$ of a pixel can be computed by the vertical and horizontal gradients of the depth map as} 
\begin{equation}
    n^{g}_{i} = [-\nabla_{x}(d_{i}), -\nabla_{y}(d_{i}), 1]^\top.
\end{equation}

\noindent The total training loss is the sum of the three losses as

\begin{equation}
\begin{aligned}
    L = L_{1} + L_{grad} + L_{normal}. \\
\end{aligned}
\end{equation}

\section{Experiments}
\label{experiments}

In this section, the experimental setup along with the proof-of-concept experiment is introduced.


\subsection{Evaluation Criteria}

\noindent \textbf{Standard metrics.} We follow the standard MDP metrics as introduced in~\cite{eigen2014} and report results as mean absolute relative error ($absrel$), root mean squared error ($rmse$), and the accuracy under threshold ($\delta_{i}<1.25^{i} = 1,2,3$). 

\noindent \textbf{Occlusion boundaries.} \evin{Since standard metrics evaluate depth distances in an averaged manner, Koch \etal \cite{koch2019} proposed another set of metrics to evaluate finer details in terms of occlusion boundaries and planarity. The Depth Boundary Error (DBE) calculates the accuracy ($\epsilon_{acc}$) and completeness ($\epsilon_{com}$) of occlusion boundaries by comparing predicted depth map edges to the annotated map of occlusion boundaries. The Truncated Chamfer Distance (TCD) is then calculated as:}

\begin{equation}
    \epsilon^{acc}_{DBE} = \frac{1}{\sum_{i}{Y_{i}}} \sum_{i} {E_{i}} \cdot Y_{i},
\end{equation}

\noindent \evin{where $Y_{i}$ is the distance between the $i$'s predicted depth map's edge pixel, extracted by Canny edge detector, to the corresponding nearest ground truth edge pixel. If this distance is greater than 10 pixels, $E_{i}$ amounts to zero to reject mismatched pixels. Oppositely, completeness  $\epsilon^{com}_{DBE}$ is calculated by calculating the distances from ground truth edges to prediction to measure how much of the existing boundaries are measured. Furthermore, Directed Depth Error (DDE) assesses whether the overall prediction falls too short or too far, by calculating the percentage of depth values that lie behind ($\epsilon^{-}$), in front ($\epsilon^{+}$), and in the proximity ($\epsilon^{0}$) of a predefined median plane. Finally, the Planarity Error (PE) denotes the surface normal error on planar region maps computed as 3D shift distance ($\epsilon^{plan}$) and angular difference ($\epsilon^{orie}$). As we focus on occlusion boundary quality, we mainly consider DBE along with the standard metrics. We state the results for PE and DDE for completeness, and refer the reader to \cite{koch2019} for further details on their formulation.}

\subsection{Datasets}
\noindent \textbf{NYU v2.} It consists of images collected in real indoor scenes \cite{silberman2012}. Depth is captured with a Microsoft Kinect camera. The raw dataset of RGB depth pairs (approximately 120K images) has no semantic labels. The authors created a smaller split with semantic and instance labels, along with the refined depths and normals. In our experiments, we refer to this smaller split, which contains 1449 images in total, namely 795 for training and 654 for testing. 

\noindent \textbf{NYU v2 - OB.} Occlusion boundary annotations on the NYU v2 test data for evaluation purposes is released by Ramamonjisoa \etal~\cite{ramamon2020} within this dataset. We use it to evaluate the occlusion boundaries of our depth predictions.

\noindent \textbf{iBims.} This dataset~\cite{koch2019} is presented as a set of around 100 images (for evaluation only) along with novel metrics on occlusion boundaries and planarity scores. They provide rich annotations of dense depth maps from different scenes, with occlusion boundaries and planar regions. 

\noindent \evin{\textbf{KITTI.} Finally, this dataset provides a monocular depth prediction benchmark~\cite{kitti}. There are no occlusion boundary annotations given, so it is not possible to evaluate DBE scores. Hence, we only provide quantitative results \textit{w.r.t} the standard metrics and employ qualitative samples to show the effectiveness of our method on outdoor scenes. In the following experiments, we trained models with 5000 sample images and took the original test split from KITTI-Eigen \cite{kitti} for testing (629 images). }

\subsection{Overfitting Experiment}

To test the effectiveness of our Instance Convolutions (IC), we first conduct overfitting experiments, comparing classical convolution based MDP methods against their IC counterpart. \evin{An exemplary outcome for this experiment is presented in Fig.~\ref{fig:intro}. Here, the classical convolutions achieve an $absrel$ of 0.009 and DBE $\epsilon_{acc}$ of 0.52, whereas our proposed IC reports 0.007 and 0.44 for $absrel$ and DBE $\epsilon_{acc}$. 
This confirms our expectation that IC has the capacity to prevent the boundary smoothing issue of standard convolutions, by only considering the relevant pixels within the segments.
}


\begin{table}[t!]
    \caption[NYUv2 state-of-the-art comparison]
  {NYU v2 Test Split state-of-the-art comparison for with and without Instance Convolution.
  } \label{tab:nyu_sotacomp}
  \centering
  \scalebox{0.74}{
      \begin{tabular}{ l |  l l | l l l | l l}
        \toprule
          \multicolumn{1}{c}{\bfseries Method} & \multicolumn{2}{c}{\bfseries Error $\downarrow$} & \multicolumn{3}{c}{\bfseries Accuracy $\uparrow$} & \multicolumn{2}{c}{\bfseries DBE $\downarrow$} \\
           & $absrel$ & $rmse$ & $\delta_{1}$ & $\delta_{2}$ & $\delta_{3}$ & $\epsilon_{acc}$ & $\epsilon_{com}$ \\
        \midrule
          SharpNet \cite{ramamonjisoa2019sharpnet} & \textbf{0.116} & \textbf{0.448}  & \textbf{0.853} & 0.970 & 0.993 & 3.041 & 8.692 \\
          + Instance Conv.  & 0.124 & 0.456 & 0.847 & \textbf{0.971} & \textbf{0.993} & \textbf{1.961} & \textbf{6.489} \\
          \midrule
          VNL \cite{yin2019virtualnormal} & \textbf{0.112} & \textbf{0.417} & \textbf{0.880} & \textbf{0.975} & \textbf{0.994} & 1.854 & 7.188\\
          + Instance Conv. & 0.117 & 0.425 & 0.863 & 0.970 & 0.991 & \textbf{1.780} & \textbf{6.059} \\
          \midrule
          BTS \cite{lee2019big} & \textbf{0.110} & \textbf{0.392} & \textbf{0.885} & \textbf{0.978} & \textbf{0.994} & 2.090 & \textbf{5.820} \\ 
          + Instance Conv. & 0.121 & 0.467 & 0.848 & 0.964 & 0.993 & \textbf{1.817} & 6.197 \\
        \bottomrule  
      \end{tabular}
  }
  
\end{table}

\begin{figure}[t]
\centering
\includegraphics[width=0.95\linewidth]{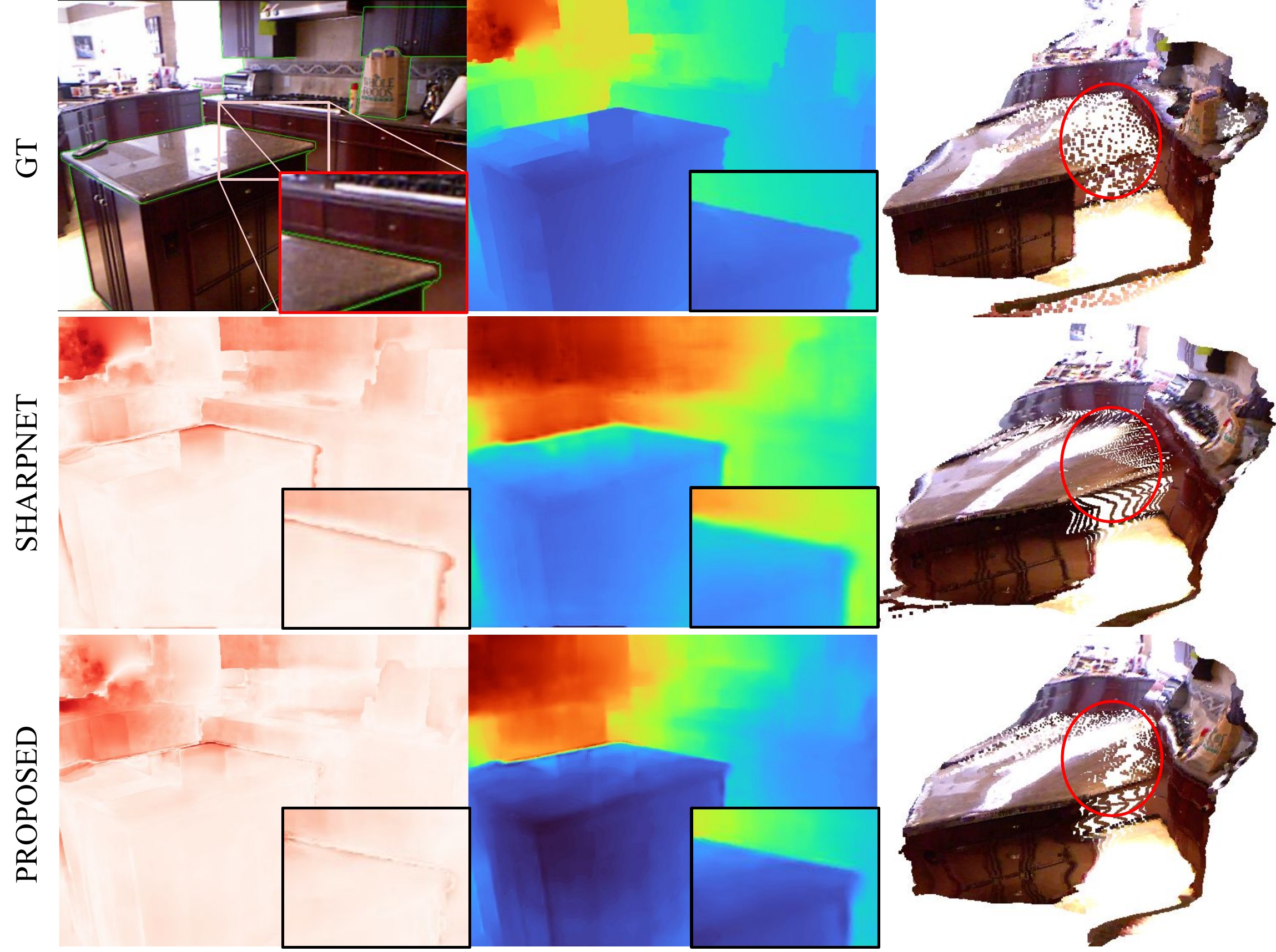}
\caption{\textbf{Error map comparison in NYUv2}, (top) the ground truth, (center) the SharpNet baseline \cite{ramamonjisoa2019sharpnet}, (bottom) the proposed IC model. The results are given as error maps, depth maps and point clouds. The improvement on the occlusion boundary is best visible when zoomed in.
}
\label{fig:error_map}
\end{figure}

\begin{table*}[htpb!]
    \caption{Quantitative results on iBims for standard metrics and PE, DBE, and DDE metrics  for different MDP methods.}
	\label{tab:ibims_results}
	\begin{center}
  \resizebox{0.85\columnwidth}{!}{
  \begin{adjustbox}{max width=\textwidth}
	\begin{tabular}{l | l l l l l l | l l | l l | l l l}
		\toprule
		
		\multicolumn{1}{c}{\bfseries Method} & \multicolumn{3}{c}{\bfseries Error  $\downarrow$} & \multicolumn{3}{c}{\bfseries Accuracy  $\uparrow$} & \multicolumn{2}{l}{\bfseries \footnotesize PE  (in $cm/\degree$)  $\downarrow$} &\multicolumn{2}{c}{\bfseries \footnotesize DBE (in $px$)  $\downarrow$} & \multicolumn{3}{c}{\bfseries \footnotesize DDE (in $\%$)}  \\

		\cmidrule(r){2-7} \cmidrule(r){8-9} \cmidrule(r){10-11} \cmidrule(){12-14}

		& \multicolumn{1}{l}{$absrel$} &
		\multicolumn{1}{l}{$log_{10}$} &
		\multicolumn{1}{l}{$rmse$} &
		\multicolumn{1}{l}{$\delta_{1}$} &
		\multicolumn{1}{l}{$\delta_{2}$} &
		\multicolumn{1}{l}{$\delta_{3}$} &
		\multicolumn{1}{l}{$\epsilon^{\text{plan}}$} &
		\multicolumn{1}{l}{$\epsilon^{\text{orie}}$} &
		\multicolumn{1}{l}{$\epsilon^{\text{acc}}$} &
		\multicolumn{1}{l}{$\epsilon^{\text{comp}}$} &
		\multicolumn{1}{l}{$\epsilon^0$ $\uparrow$} & 
		\multicolumn{1}{l}{$\epsilon^-$ $\downarrow$} &
		\multicolumn{1}{l}{$\epsilon^+$ $\downarrow$} \\

		\cmidrule(r){1-1} \cmidrule(r){2-2} \cmidrule(r){3-3} \cmidrule(r){4-4} \cmidrule(r){5-5} \cmidrule(r){6-6} \cmidrule(r){7-7} \cmidrule(r){8-8} \cmidrule(r){9-9} \cmidrule(r){10-10} \cmidrule(r){11-11} \cmidrule(r){12-12} \cmidrule(r){13-13} \cmidrule(){14-14}

		Eigen \cite{eigen2014}                 & 0.32 & 0.17 & 1.55 & 0.36 & 0.65 & 0.84 & 7.70 & 24.91 & 9.97 & 9.99 & 70.37 & 27.42 & 2.22 \\
		Laina \cite{laina2016}                 & 0.26 & 0.13 & 1.20 & 0.50 & 0.78 & 0.91 & 6.46 & 19.13 & 6.19 & 9.17 & 81.02 & 17.01 & 1.97 \\
		Liu \cite{liu2015}                   & 0.30 & 0.13 & 1.26 & 0.48 & 0.78 & 0.91 & 8.45 & 28.69 & \textbf{2.42} & 7.11 & 79.70 & 14.16 & 6.14 \\
		Li  \cite{li2017}                  & \textbf{0.22} & 0.11 & 1.09 & 0.58 & 0.85 & \textbf{0.94} & 7.82 & 22.20 & 3.90 & 8.17 & 83.71 & 13.20 & 3.09 \\
		Liu  \cite{liu2018planenet}                  & 0.29 & 0.17 & 1.45 & 0.41 & 0.70 & 0.86 & 7.26 & 17.24 & 4.84 & 8.86 & 71.24 & 28.36 & \textbf{0.40} 
		\\
		\hline
		SharpNet \cite{ramamonjisoa2019sharpnet}	 & 0.26 & 0.11 & 1.07 & \textbf{0.59} & \textbf{0.84} & \textbf{0.94} & 9.95 & 25.67 & 3.52 & 7.61 & \textbf{84.03} & 9.48 & 6.49
		\\
		with Instance Conv. & 0.29 & 0.12 & 1.14 & 0.55 & 0.82 & 0.92 & 9.83 & 25.88 & 3.11 & 7.83 & 81.84 & \textbf{8.27} & 9.88 \\

        \hline
		BTS \cite{lee2019big} & 0.24 & 0.12 & 1.08 & 0.53 & 0.84 & 0.94 & 7.24 & 20.51 & 2.50 & 5.81 & 82.24 & 15.50 & 2.27 \\
		with Instance Conv. & \textbf{0.22} & 0.11 & 1.11 & 0.57 & \textbf{0.86} & \textbf{0.94} & 6.76 & 19.39 & 3.71 & 8.01 & \textbf{84.04} & 13.3 & 2.67 \\
		\hline
		VNL \cite{yin2019virtualnormal} & 0.24 & 0.11 & \textbf{1.06} & 0.54 & 0.84 & 0.93 & 5.73 & 16.91 & 3.64 & 7.06 & 82.72 & 13.91 & 3.36 \\
		with Instance Conv. & 0.23 & \textbf{0.10} & \textbf{1.06} & 0.58 & 0.85 & 0.93 & \textbf{5.62} & \textbf{16.53} & 3.03 & 7.68 & 83.85 & 13.26 & 2.87 \\
		\bottomrule
	\end{tabular} 
	\end{adjustbox}
	}
	\end{center}
\end{table*}

\subsection{Comparison to State-of-The-Art}

In Table~\ref{tab:nyu_sotacomp}, we compare our results from NYU v2 with three state-of-the-art approaches, namely SharpNet \cite{ramamonjisoa2019sharpnet}, VNL \cite{yin2019virtualnormal} and BTS \cite{lee2019big}. Thereby, the proposed architecture (Fig. \ref{fig:architecture}) uses these pre-trained models for latent depth feature extraction and applies Instance Convolution based blocks.

All models are trained using PyTorch on a NVidia RTX 2080Ti for 35 epochs with Adam optimizer, started learning rate with $1e-4$ and decreased by $0.1$ every 10 epochs. The loss terms in Eq. 6 have equal weights of 1. We use SLIC \cite{achanta2012slic} to obtain super-pixels with 64 segments and set sigma to 1. \evin{SharpNet \cite{ramamonjisoa2019sharpnet} is trained with a batch size of 4, extracting the feature embedding from \texttt{UpConv1} layer having a shape of $128 \times 120 \times 160$. For BTS \cite{lee2019big} we use a batch size of 3, and forward the output of the \texttt{iConv1} feature layer of size $32 \times 480 \times 640$. Finally, VNL \cite{yin2019virtualnormal} also utilizes a batch size of 3, and propagates the feature from the final layer of shape $256 \times 480 \times 640$.} Our proposed model then appends 3 layers of Instance Convolutions with gradually decreasing number of feature channels. The feature map resolution remains constant with a final kernel of size $1\times1$.

\begin{figure}[t!]
\centering

\includegraphics[width=0.97\textwidth]{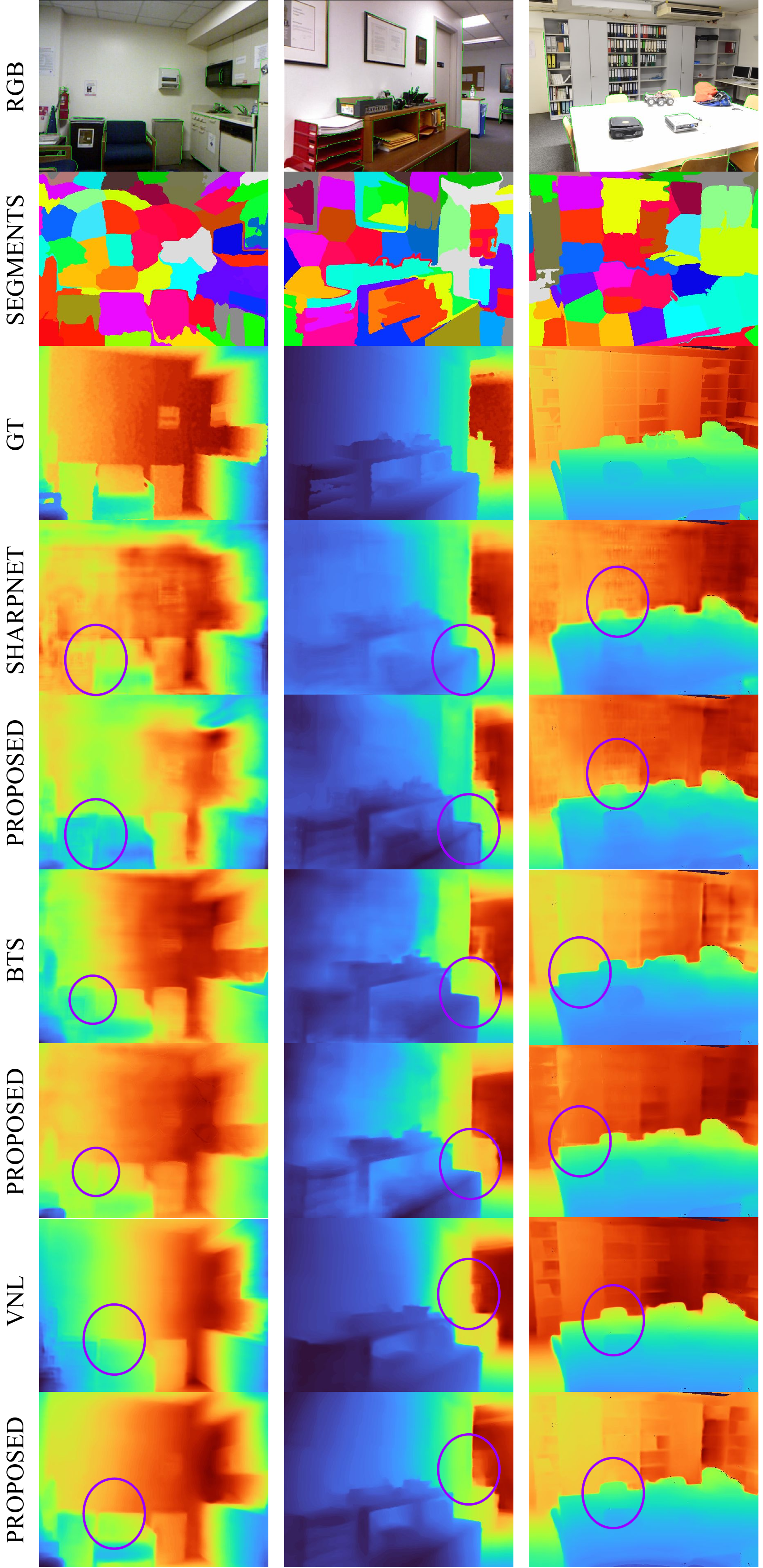}

\caption{\textbf{Qualitative results.} The first two columns are samples from NYUv2 \evin{and last column from iBims dataset}, rows respectively show the RGB input image, SLIC output for the input image, ground truth depth, and depth maps from original SharpNet \cite{ramamonjisoa2019sharpnet}, BTS \cite{lee2019big} and VNL \cite{yin2019virtualnormal} models, and the improved depth maps from proposed Instance Convolution models.}
\label{fig:qualitative_res}
\end{figure}

We outperform the original methods with respect to the occlusion boundary metrics, and report comparable results for classical metrics. The qualitative results in the Fig.~\ref{fig:qualitative_res} further supports these findings, where the proposed Instance Convolution predictions of each model have sharper occlusion boundaries, and resulting depth maps. Further, in Fig. \ref{fig:error_map} we show an additional qualitative example to highlight the error distribution in a depth map. We observe that the error map of the SharpNet baseline (center row) has higher values along the occlusion boundaries, while ours (bottom row) are mostly uniform (\emph{c.f.} zoomed-in region).

In Table~\ref{tab:ibims_results}, we further report our results with respect to the iBims evaluation dataset, in order to assess the generalizability of our method. Note that this dataset is used only for inference (\ie no training), to measure whether the models are capable of detecting depth values. Here, \evin{in most cases} our models again improved the counterpart depth models in terms of DBE accuracy and completion metrics. \evin{We additionally provide the qualitative results for iBims samples in Fig. \ref{fig:qualitative_res} third column, with the highlighted occlusion boundary regions.} 

Notably, on iBims, the model with the best absrel value does not have the best DBE score (Li \etal~\cite{li2017} 0.22 absrel, 3.90 DBE \textit{vs.} Liu \etal~\cite{liu2015} 0.30 absrel, 2.42 DBE). \evin{This can also be seen on NYU results, where our Instance Convolution improving DBE scores, also the occlusion boundaries as seen in qualitative illustrations, however being on-par on absrel metrics. This conceptually agrees with the fact that absrel averages out per pixel distances, while DBE calculates the drift errors on occlusion maps. This could explain the degradation on SharpNet absrel score on iBims.}

\begin{figure}[t!]
\begin{center}
\includegraphics[width=0.9\linewidth]{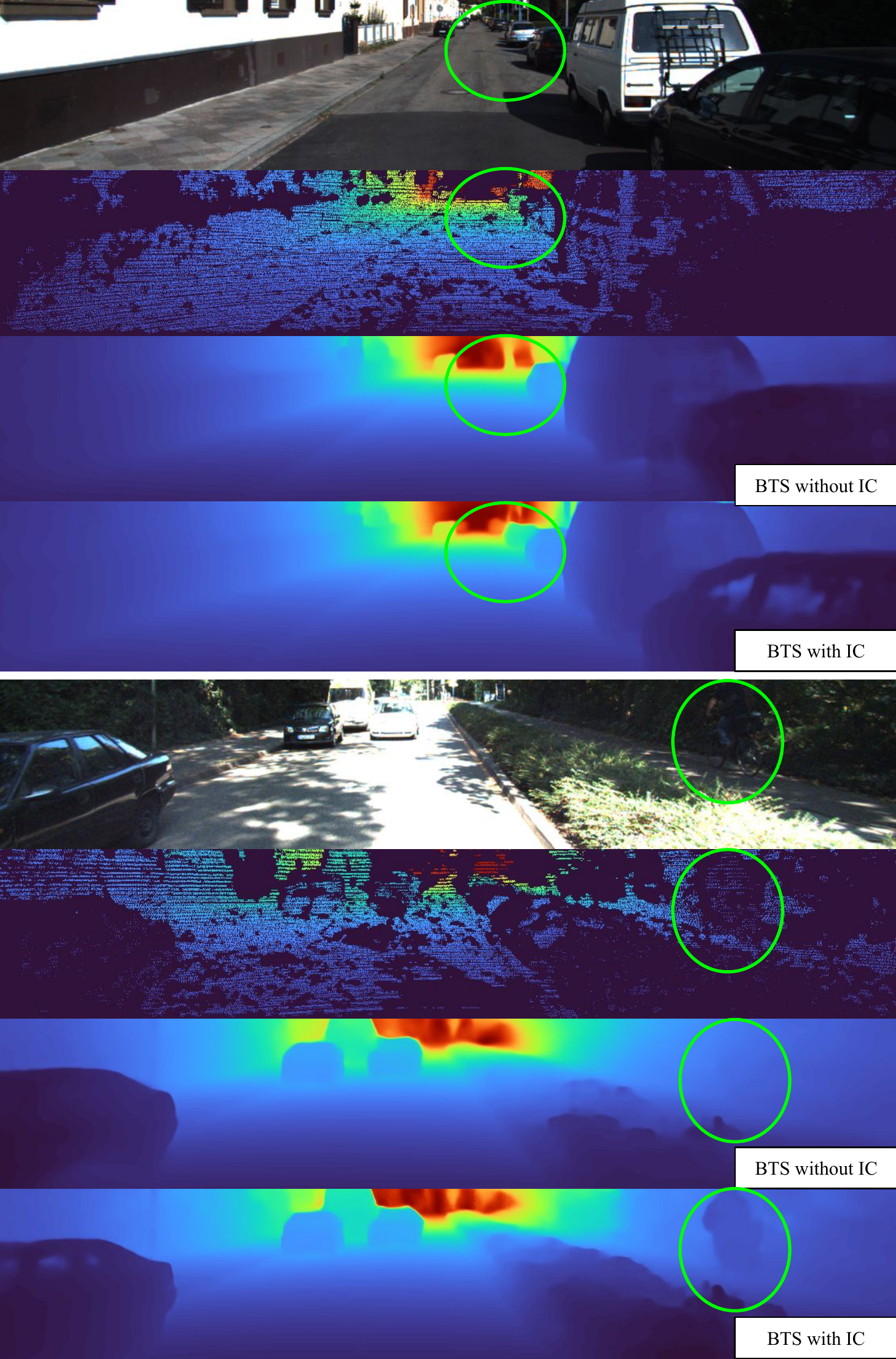}
\end{center}
\caption{\evin{\textbf{Qualitative results on KITTI \cite{kitti}}, using the BTS \cite{lee2019big} backbone, showing (from top to bottom) the input RGB, ground truth depth map, BTS prediction, and proposed IC model prediction. The improvements on occlusion boundaries around objects are highlighted with green-circles.}}
\label{fig:kitti}
\end{figure} 

\evin{Finally, we evaluated our model on KITTI~\cite{kitti}, extending BTS with a DenseNet-161 backbone \cite{densenet}. Since KITTI does not provide annotations for occlusion boundaries, we mostly rely on qualitative evaluation to assess the performance in terms of boundary quality. Fig.~\ref{fig:kitti} illustrates a few examples, comparing the BTS baseline with ours. Notably, our method not only produces sharper edges for the objects, but it is also able to recover sharp contours for the biker in the second image, which is completely missed by standard BTS, thanks to our over-segmentation formulation. Further, we are capable of producing sharper results without any loss in quality. In particular, using the standard metrics BTS reports an error of 0.062 and 2.798 for absrel and rmse, respectively, while we instead report an error of 0.06 and 2.479.}

\subsection{Ablation Studies} \label{sec:ablation}
\evin{In Table~\ref{tab:ablation_table}, we present an ablation for different parameters and configurations using SharpNet \cite{ramamonjisoa2019sharpnet} as backbone. Instance convolution based models are shown in the table with IC or Standard Convolutions as SC, following by the number of segmented regions. The baseline model is the original SharpNet model without SC/IC.}

\noindent \textbf{Super-pixels information.} We trained a model with SC and provided the super-pixel segmentation map as an extra input to each convolutional layer. As shown in Table III line SC 64, DBE score is better than the baseline, however, worse than the proposed IC. It shows that the proposed IC method is indeed able to use the super-pixels information as desired to reason on occlusion boundaries. 

\noindent \textbf{Number of segments in SLIC.} We ablate different number of super-pixel segments. As expected, increasing number of segments improve DBE accuracy, yet, induce a little loss in absrel. Notice that this is also the case for most sota works. Best performing works for edges are often worse on absrel, which could be caused by imperfect annotations.

\noindent \textbf{Over segmentation with BASS.} We also qualitatively evaluated our methods using BASS~\cite{uziel2019bayesian} to extract super-pixels. As shown in the Fig.~\ref{fig:over_seg}, BASS is able to retrieve more detailed segments from the image, however it also detects overly noisy edges due to redundant number of segments (200-250), which increases the model complexity and makes the learning more difficult.

\noindent \textbf{Instance masks.} To compare the quality of instance mask prediction to unsupervised segmentation, we ablated our method with the state-of-the-art instance mask prediction method PointRend \cite{kirillov2020pointrend}. As both the absrel and the DBE results were poorer than the baseline, this proved the effectiveness of over-segmentation method, most likely because of detecting self-occlusions within the images. \evin{Additionally, we show our results using ground truth instance masks, which, as expected, improve performance over predicted masks. Nevertheless, very interestingly, our proposed IC approach with 128 segments performs a little better than ground truth masks \textit{w.r.t.} $\epsilon_{acc}$, which we contribute to the fact the instance masks neglect intra-shape boundaries (see Fig.~\ref{fig:over_seg}).}

\noindent \textbf{Runtime analysis} Full inference times are given in Table \ref{tab:ablation_table} as Frames per Second (FPS) excluding segmentation calculation, FPS* including. Each runtime is averaged on 1000 inferences. IC does not excessively alter the FPS compared to SharpNet.

\begin{table}[t!]
\caption[Ablation on the segmentation method.]
  {\evin{Ablation study on NYUv2, comparing usage of different masks (ground truth, PointRend, and BASS), super-pixels with Standard Convolutions (SC), and different number of segments (16-32-64-128) with Instance Convolutions (IC)}. } \label{tab:ablation_table}
  \centering
  \resizebox{0.85\columnwidth}{!}{
  \begin{tabular}{ l |  l l | l l | l  l}
    \toprule
      {\bfseries Method} & \multicolumn{2}{c}{\bfseries Error $\downarrow$} & \multicolumn{2}{c}{\bfseries DBE $\downarrow$} & \multicolumn{2}{c}{\bfseries Runtime} \\
        & $absrel$ & $rmse$ & $\epsilon_{acc}$ & $\epsilon_{com}$ & FPS & FPS* \\
     \midrule
     SharpNet~\cite{ramamonjisoa2019sharpnet} & \textbf{0.12} & \textbf{0.45} & 3.04 & 8.69 & \textbf{16.7} & \textbf{16.7}\\
    \midrule
      \evin{GT Masks} & \evin{\textbf{0.12}} & \evin{0.46} & \evin{2.05} & \evin{6.49} & \evin{13.5} &  \evin{13.5} \\
      PointRend \cite{kirillov2020pointrend} & 0.13 & \textbf{0.45} & 2.21 & 6.76 & \evin{13.5} &  \evin{3.64} \\
      \midrule
      BASS \cite{uziel2019bayesian} & \textbf{0.12} & 0.46 & 2.19 & 6.63 & \evin{13.2} & \evin{0.59} \\ 
      IC 16 & 0.14 & 0.47 & 2.07 & 6.59 & 13.5 & 3.08 \\
      IC 32  & 0.14 & 0.47 & 2.09 & 6.66 & 13.6 & 3.04 \\
      IC 64  & \textbf{0.12} & 0.46 & 1.96 & \textbf{6.48} & 13.4 & 2.97\\
      SC 64  & \textbf{0.12} & \textbf{0.45} & 2.18 & 6.63 & 15.2 & 3.05 \\
      IC 128 & 0.13 & 0.46 & \textbf{1.92} & 6.57 & 13.3 & 2.89 \\
    \bottomrule 
    
  \end{tabular} }

\end{table} 

\section{CONCLUSIONS}
In this work, we introduce a novel depth estimation method, which is particularly tailored towards tackling the problem of depth smoothing at object-discontinuities. To this end, we propose a novel convolutional operator, which avoids feature aggregation across discontinuities by means of super-pixels. Our exhaustive evaluation on NYU depth v2, iBims as well as KITTI demonstrates that the proposed method is capable of improving depth inference around edges, while almost completely maintaining the quality on the remaining regions. In the future, we want to explore how Instance Convolution can be incorporated into other domains such as semantic segmentation to similarly improve sharpness.

\bibliography{egbib}

\end{document}